\documentclass[conference]{IEEEtran}
\usepackage{booktabs}
\usepackage{graphicx}
\usepackage{hyperref}
\usepackage{enumitem}
\usepackage{amsmath}  % or \usepackage{amsfonts}
\usepackage{amssymb}

\ifCLASSINFOpdf
   \graphicspath{ {Images} }
   \DeclareGraphicsExtensions{.pdf,.jpeg,.png}
\else
\fi

\usepackage{tikz}

\usepackage{amsmath}

\begin{document}
\renewcommand\IEEEkeywordsname{Keywords}
% paper title
% can use linebreaks \\ within to get better formatting as desired
\title{Underwater Object Detection Enhancement via Channel Stabilization }

% author names and affiliations
% use a multiple column layout for up to three different
% affiliations
\author{\IEEEauthorblockN{Muhammad Ali}
\IEEEauthorblockA{Department of Machine Learning\\
Mohamed bin Zayed University of AI\\
Abu Dhabi, UAE \\
Email: muhammad.ali@mbzuai.ac.ae}
% \and
% \IEEEauthorblockN{Omar Alsuwaidi}
% \IEEEauthorblockA{Department of Computer Vision\\
% Mohamed bin Zayed University of AI\\
% Abu Dhabi, UAE \\
% Email: omar.alsuwaidi@mbzuai.ac.ae}
 \and
 \IEEEauthorblockN{Salman Khan}
 \IEEEauthorblockA{Department of Computer Vision\\
 Mohamed bin Zayed University of AI\\
 Abu Dhabi, UAE \\
 Email: salman.khan@mbzuai.ac.ae\\
 }}

% conference papers do not typically use \thanks and this command
% is locked out in conference mode. If really needed, such as for
% the acknowledgment of grants, issue a \IEEEoverridecommandlockouts
% after \documentclass

% for over three affiliations, or if they all won't fit within the width
% of the page, use this alternative format:
% 
%\author{\IEEEauthorblockN{Muhammad Ali\IEEEauthorrefmark{1},
%Homer Simpson\IEEEauthorrefmark{2},
%James Kirk\IEEEauthorrefmark{3}, 
%Montgomery Scott\IEEEauthorrefmark{3} and
%Eldon Tyrell\IEEEauthorrefmark{4}}
%\IEEEauthorblockA{\IEEEauthorrefmark{1}School of Electrical and Computer Engineering\\
%Georgia Institute of Technology,
%Atlanta, Georgia 30332--0250\\ Email: see http://www.michaelshell.org/contact.html}
%\IEEEauthorblockA{\IEEEauthorrefmark{2}Twentieth Century Fox, Springfield, USA\\
%Email: homer@thesimpsons.com}
%\IEEEauthorblockA{\IEEEauthorrefmark{3}Starfleet Academy, San Francisco, California 96678-2391\\
%Telephone: (800) 555--1212, Fax: (888) 555--1212}
%\IEEEauthorblockA{\IEEEauthorrefmark{4}Tyrell Inc., 123 Replicant Street, Los Angeles, California 90210--4321}}

% use for special paper notices
%\IEEEspecialpapernotice{(Invited Paper)}

% make the title area
\maketitle

\begin{abstract}
The complex marine environment exacerbates the challenges of object detection manifold. With the advent of the modern era, marine trash presents a danger to the aquatic ecosystem, and it has always been challenging to address this issue with a complete grip. Therefore, there is a significant need to precisely detect marine deposits and locate them accurately in challenging aquatic surroundings. To ensure the safety of the marine environment caused by waste, the deployment of underwater object detection is a crucial tool to mitigate the harm of such waste. Our work explains the image enhancement strategies used and experiments exploring the best detection obtained after applying these methods. Specifically, we evaluate Detectron 2’s backbone performance using different base models and configurations for the underwater detection task. We first propose a novel channel stabilization technique on top of a simplified image enhancement model to help reduce haze and color cast in training images. The proposed procedure shows improved results on multi-scaled objects present in the data set. Following image processing, we investigate various backbones in Detectron2 to provide the best detection accuracy for these images. In addition, we use a sharpening filter with augmentation techniques. This highlights the profile of the object, which helps us easily recognize it. We demonstrate our results by verifying them on the TrashCan Data set, both instance and material versions. We then explore the best-performing backbone method for this setting. We apply our channel stabilization and augmentation methods to the best-performing technique. We also compare our detection results from Detectron2 using the best backbones with those from Deformable Transformer. The detection result for small-sized objects in the instance-version of TrashCan 1.0 gives us a 9.53\% absolute increase in average precision, while for the bounding box, we get an absolute gain of 7\% compared to the baseline. The code will be available on \textbf{Code: }\href{https://github.com/aliman80/Underwater-Object-Detection-via-Channel-Stablization}{https://github.com/aliman80/Underwater-Object-Detection-via-Channel-Stablization}
\end{abstract}
% IEEEtran.cls defaults to using nonbold math in the Abstract.
% This preserves the distinction between vectors and scalars. However,
% if the conference you are submitting to favors bold math in the abstract,
% then you can use LaTeX's standard command \boldmath at the very start
% of the abstract to achieve this. Many IEEE journals/conferences frown on
% math in the abstract anyway.

% no keywords

% For peer review papers, you can put extra information on the cover
% page as needed:
% \ifCLASSOPTIONpeerreview
% \begin{center} \bfseries EDICS Category: 3-BBND \end{center}
% \fi
%
% For peerreview papers, this IEEEtran command inserts a page break and
% creates the second title. It will be ignored for other modes.
\IEEEpeerreviewmaketitle

\section{Introduction}
% no \IEEEPARstart
With the advent of the modern era, we face massive plastic production, less recycling, and poor waste management. Between 4 and 12 million metric tons of plastic enter the ocean each year \cite{7Society}, and this number is extrapolated in the future to be at least thrice of the current amount within twenty years \cite{7Society}. Such inadequate handling of plastics negatively impacts marine life like sea turtles, whales, seabirds, fish, coral reefs, and countless other marine species and habitats. Scientists estimate that more than half of the world’s sea turtles and nearly every seabird on Earth have eaten plastic in their lifetimes. This plastic and other waste pollution harm human life, damaging beautiful beaches, coastlines, and others.
Furthermore, trash deposits in aquatic environments destroy marine ecosystems and constitute a continuing economic and environmental threat. Therefore it is worth working in this domain to apply AI models for better detection \cite{7Society}. Different environmental and government agencies have tried multiple methods to get rid of this underwater waste, with few being very successful \cite{Hong2020TRASHCAN:DEBRIS}. Additionally, while trying all these methods, we need to take care of the underwater ecosystem, making the task harder.
Researchers work to remove waste from the ocean surface by using LIDAR to trace trash on beaches, using sonar imagery to detect underwater debris \cite{Ge2016Semi-automaticOPEN}. Data produced by generative models, also improves  detection algorithms for trash detection  \cite{Hong2020TRASHCAN:DEBRIS}.\\
Underwater object detection is inspired by general object detection methods \cite{LiWaterGAN:Images}, and underwater image enhancement techniques \cite{PhamRoadR-CNN}. 
Wei-Hong et al. \cite{LinROIMIX:DETECTION} presents a model with better generalization capability by using  augmentation schemes in addition to ROIMix. To enhance the domain diversity for different datasets, Hong Liu et al. \cite{LiuWQTDETECTION} uses the data augmentation method. For small-scale underwater detection, Long Chen et al. \cite{Chen2020UnderwaterLearning} presents the Sample-WeIghted hyPEr Network (SWIPENet) but suffers from time issues as it is an ensemble of many deep learning methods. To produce a realistic training set, Kaiiming Hee et al. \cite{He2011SinglePrior} propose a method to remove haze from a single image  by relating to the statistics of haze-free outdoor images. Despite exciting research done in underwater image enhancement, we motivate ourselves to use it for waste detection using our simple image enhancement model with Detectron2. 
In this paper, we explore  object detection performance shown by \cite{Facebookresearch/detectron2:Tasks.} with various backbones for underwater image detection enhancement and shows improvement in accuracy using our novel image enhancement strategies and augmentations techniques which help us achieve good performance. To be more specific our contributions include following:
%%

%\begin{enumerate}

\begin{itemize}[noitemsep,topsep=0pt]
\item We take the input image and use LAB-Stretching and global stretching inspired by  \cite {Huang2018Shallow-waterAcquisition}
to pre-process the image for removing colour haze and cast in images. 
\item  We then  propose the Channel Stabilization module which helps to reduce  dominance of one specific colour in the underwater imagery.
\item We use augmentation techniques along with  with sharpening filter to highlight the  contours, though it affects the qualitative display.
\item We explore the best performing back bones for object detection for this processed image. After extensive set of experiments we observe that for underwater images in low waters, Retina Net exceeds others in performance.
   % \item All these detections are tried using pre trained weights with various backbones as well as the results are also confirmed by using deformable transformer for detection.
     %The cross domain results are verified by checking the performance on UAV-dataset as well as on TACO data set.
     \end{itemize}
%\end{enumerate}	
%%
The rest of the paper is organized as follows: Section 2 describes the related work in the domain, which provides relevant context and motivation for the implementation of our approach. Section 3 explains the overall methodology used; it describes the image enhancement model (IEM) along with channel sharpening and stabilization modules. Experimentation, results, and evaluations are given in Section 4. Finally, Section 5 concludes the paper with findings and future research directions.

\section{Related Work}
To process the underwater images there are two categories of algorithms and techniques: physics-based  and image-based. Image-based methods provide a much simpler and straightforward solution. Color equalization methods are used to remove single channel (color) dominance and deal with the color cast problem of underwater images, where the blue or green color is dominant, whereas, in our proposed solution, we suggest a color stabilization module. Iqbal et al. \cite{IqbalUnderwaterModel} propose integrated color model (ICM) as well as  unsupervised color correction method (UCM) \cite{Iqbal2010EnhancingMethod}. To improve the contrast and color cast, they use histogram stretching in RGB color mode and saturation-intensity stretching in the HSI color model.\\
In one work, they use a dark channel prior to estimate and remove the haze from an image \cite{He2011SinglePrior}. Another work in \cite{Peng2017UnderwaterAbsorption} replaces the dark channel prior with an image blurriness map and gets better results. Work in  \cite{Li2020AnBeyond} uses white balance, histogram equalization, and Gamma correction. This method shows good results in diverse conditions. In another work \cite{DatPham2019ClassifierModel} use GAN based approach. Furthermore, Scientist also use a denoising image transformer to remove distortion from underwater images \cite{Tian2020Attention-guidedDenoising}.\\ 
The image enhancement step reduces the challenges faced by object detectors. In general, object detection algorithms can be classified as One Stage and Two Stage detection pipelines. In the two-stage approach, like R-CNN \cite{RichXplore} generates ROIs, input these to neural network which extracts high-level features from these regions and using these to detect unique objects. Faster R-CNN \cite{PhamRoadR-CNN} put forward an efficient and accurate network to achieve combined end-to-end object detection nearly.
On the other hand, the one-stage approach gives higher efficiency and easier model training compared to the two-stage ones and give competitive performance \cite{RichXplore}. Retina Net is a one-stage object detection model that utilizes a focal loss function to address class imbalance during training \cite{FocalCode}. It is a type of CNN (Convolutional Neural Network) architecture that uses the Feature Pyramid Network (FPN) with ResNet. Lu tan et al. \cite{Tan2021ComparisonIdentificationb} motivates the RetinaNet usage by showing its improved performance in terms of mAP compared to single stage detector(SSD) and Faster R-CNN, but these do not use underwater images for their work.

\section{Methods}
\begin{figure*}[h]
   \centering
    \includegraphics[scale=0.30]{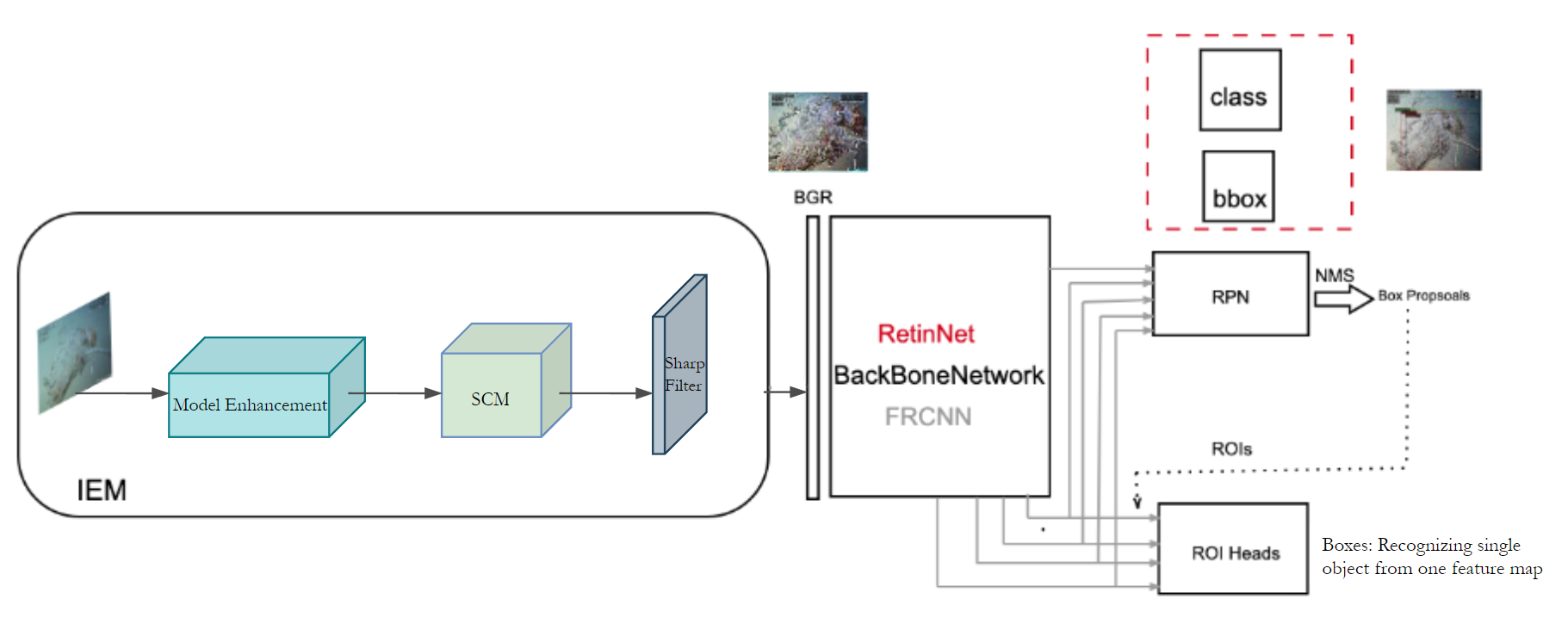} 
    \caption{The Architecture of proposed method. In the first block pre-processing is done with image enhancement model then  channel stabilization scheme is applied. Image is then  processed in Detectron2 using various backbones: RetinaNet output is given in red block while for  Faster R-CNN it follows flow given in grey and use ROI heads for final detction. }
    \label{model-updated.png}
\end{figure*}
\subsection{Overview}
CNN-based approaches perform pretty well with image enhancement methods. To the best of our knowledge, we are the first to exploit Detectron2's \cite{Facebookresearch/detectron2:Tasks.} ability to improve underwater detection using different backbones with channel stablization and enhancement.  % Very few people use Detectron2 for underwater detection, at least in the literature I have seen. 
We explore the best backbone, which gives us improved performance using our image enhancement strategy, and we find out that in post-image enhancement with single-stage detector, RetinaNet \cite{FocalCode} supercedes others in terms of accuracy.
Our architecture is composed of the following parts: (a) an input layer to
prepossess image with simplified image enhancement model (b)  channel stabilization module; (c) a Faster CNN/Retina Net backbone which extracts features by convolutions. 
Our method takes the original image and passes through Image Enhancment Module(IEM) inspired by RGHS %\cite{Huang2018Shallow-waterAcquisition} 
with a channel stabilization module, which removes the color haze to improve the visibility of the images. We then explore Detection2 for an optimized backbone to predict multi-scale classes and bounding boxes of objects as given in Figure 1. Moreover, we apply a sharpening filter to highlight the objects' contours and edges, along with data augmentation techniques to reduce the training loss and avoid over-fitting.  

%\subsubsection{Datasets} 
%Subsubsection text here.

\subsection{PreProcessing}

\subsubsection{Image Enhancement}
Underwater images  suffer from contrast, fuzzy, and color cast due to physical phenomena of scattering and absorption. In order to resolve these issues, various image enhancement and color correction models are used \cite{Huang2018Shallow-waterAcquisition}. This model works well for shallow water. For our data set, images are taken in low water conditions; therefore, inspired by this model, we modify the images such that in the color correction step, after simple global stretching to the 'L' component(brightness) of the image we  provide  adjustment  for both 'a' and 'b' components in CIE-Lab color space. We get improvement in brightness and saturation of image after application of the adaptive stretching of 'L', 'a', and 'b'. After this step we pass it through the channel stabilization module. That image processing sequence gives us quite promising results considering its simplicity and speed for real-time object detection. Figure 2 give us the overall architecture of a model where image after passing through IEM block goes to network for training and detection. In Figure 3, we show  the building blocks of the enhancement module which give us overall layout of the IEM module with different steps involved. 

\subsubsection{Channel Stabilization}
Here we propose a simple yet novel image enhancement technique to improve the existing enhancement module used in our model. Initially, we start by posing some reasonable assumptions about our training data that will help us simplify the problem of haziness and single channel (color) dominance in underwater images. Firstly, we assume that, due to the nature of our training images being underwater, the blue channel has the largest distribution of intensity values across its pixels compared to the other channels; hence our training images are blue color dominant. Secondly, on the other hand, by the same reasoning, the red channel will have the smallest distribution of intensity values as compared to the other channels. These assumptions are mathematically formulated as shown in Equation \ref{eq:1}. We argue that these assumptions are justified and legitimate due to the physical phenomena of \emph{absorption and scattering of light}; where the water absorbs colors in the red part of the light spectrum, causing an underwater object to be deprived of the red color. At the same time, water scatters the blue wavelengths of light, similar to the scattering of blue light in the sky, due to the high frequency of the blue wavelength of light, causing an underwater object to look bluish. Therefore, underwater images are blue-biased. The technique works by splitting the individual RGB channels of an input image in order to find the individual average intensity value of each channel:
% $$ 
% \bar{I_{R}} = mean(I_R), \;\;
% \bar{I_{G}} = mean(I_G), \;\;
% \bar{I_{B}} = mean(I_B),
% $$
\begin{equation}
 \bar{I_{R}} = mean(I_R), \;\;
 \bar{I_{G}} = mean(I_G), \;\;
 \bar{I_{B}} = mean(I_B)
\end{equation}
where the red channel, green channel, and the blue channel are represented by $I_R$, $I_G$, and $I_B$, respectively.
Here, we formalize our assumption regarding the pixel intensity behaviour of underwater images as:

\begin{equation} \label{eq:1}
\bar{I_{B}} \geq \bar{I_{G}} \geq \bar{I_{R}}
\end{equation}
We define the mean of the channel means $(I_{mean})$ as follows:
\begin{equation} \label{eq:2}
I_{mean} = \left(\frac{\bar{I_{R}} + \bar{I_{G}} + \bar{I_{B}}}{3}\right)
\end{equation}
Now we introduce a proportionality scale factor to refactor the initial blue-biased channels as follows:
\begin{align}
\hat{I}_{R} &= \left(\frac{I_{mean}}{\bar{I_{R}}}\right)\odot I_R \\
\hat{I}_{G} &= \left(\frac{I_{mean}}{\bar{I_{G}}}\right)\odot I_G \\
\hat{I}_{B} &= \left(\frac{I_{mean}}{\bar{I_{B}}}\right)\odot I_B
\end{align}
Based on the assumptions made in Equation \ref{eq:1}, and by utilizing the $I_{mean}$ from Equation \ref{eq:2}, we can deduce that the red channel's scaling factor will always be:
\begin{equation}
\left(\frac{I_{mean}}{\bar{I_{R}}}\right) > 1
\end{equation}
while the blue channel's scaling factor will always be:
\begin{equation}
\left(\frac{I_{mean}}{\bar{I_{B}}}\right) < 1
\end{equation}
though we cannot conclude anything regarding the green channel's scaling factor.
Hence, we adequately re-scale the image's individual channel intensity distribution to stabilize the overall color quality of the image.
Furthermore, this technique works for the general case, where if any particular color channel is dominant in any image, then that color is re-scaled down accordingly. We can the illustration of this method in Figure 3.

\begin{figure}[h]
   \centering
    \includegraphics[scale=0.18]{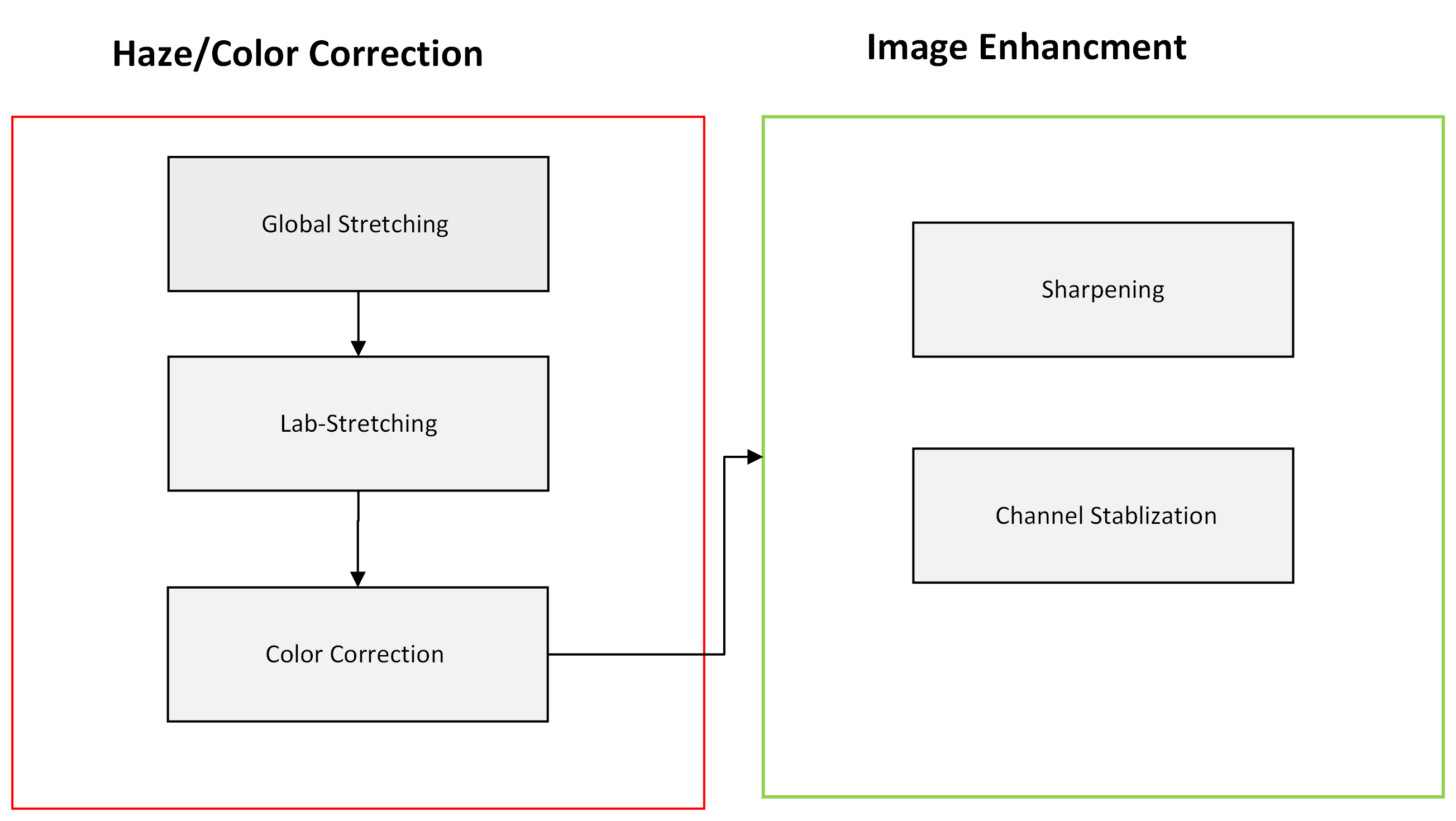} 
    \caption{The Image Enhancement block and Channel Stabilization module  which equalizes the image brightness and color restoration.   }
    \label{model-updated.png}
\end{figure}
 In this module, we use stretching to standardize the original image in such a way that the weak channel is subtracted from the image, and then it we multiply it by 255 in order to cover all the ranges of the missing colors, achieving standardization.
%One sample image after processing in this module is given in Figure 3. Though computationally less expensive it give us good result. 

\begin{figure}[h]
   \centering
    \includegraphics[scale=0.25]{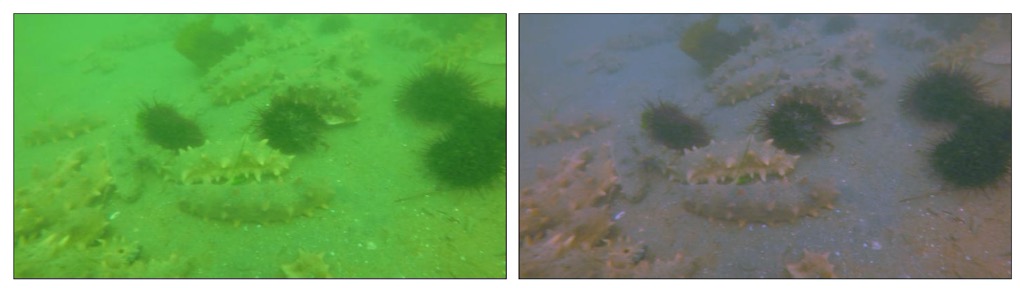} 
    \caption{Channel Stabilized image. }
    \label{fig:ch-stabilize.jpg}
\end{figure}
\subsection{Augmentations}
After the channel stabilization module, and before feeding the image to the model during training, we apply a sharpening \cite{Jiang2021UnderwaterAttention} filter to enhance the contours and visibility after the channel stabilization module and before sending the image to the model during training.\\ In order to increase accuracy and decrease over-fitting, we additionally use image cropping and horizontal flipping. We evaluate numerous approaches to strengthen the model's capacity for object identification and increase its resistance to various visual difficulties, such as occlusion, light fluctuations, and others. As seen in Fig. 8, using a sharpening filter results in decent accuracy with some observable artifacts. By further increasing this contrast, we can achieve good results because the item's contour includes high contrast information between the object and marine environment. For this reason, we use a sharpening filter after the image enhancement and before the channel stabilization scheme. Inspired by the work from \cite{Jiang2021UnderwaterAttention} In our case, we employ horizontal flip to expand the data set and get good outcomes. Applying a sharpening filter improves performance by highlighting the features of the item. We also experiment with the  "broken mirror" method of new image augmentation, which involves randomly dividing an image across its column dimension and flipping both sides of the partition vertically. To see the effects of background we use Background Removal, which to a certain extent removes the background's impact on an image. It helps in reducing the  Convolutional neural networks' inductive biases. Although visually appealing, as seen in Figure 4, it degrades the mAP score in our settings.
We try various combination of augmentation methods, though some combination degrades the performance especially in very dark or night time images. We try various augmentation method to analyze their effects in different settings, otherwise standard augmentation methods like horizontal flipping also works well.
%\end{enumerate}
%\begin{figure}[h]
  % \centering
   % \includegraphics[scale=0.20]{b-mirror.jpg} 
    %\caption{broken-mirror . }
    %\label{fig:b-mirror.jpg}
%\end{figure}

\begin{figure}[h]
   \centering
    \includegraphics[scale=0.20]{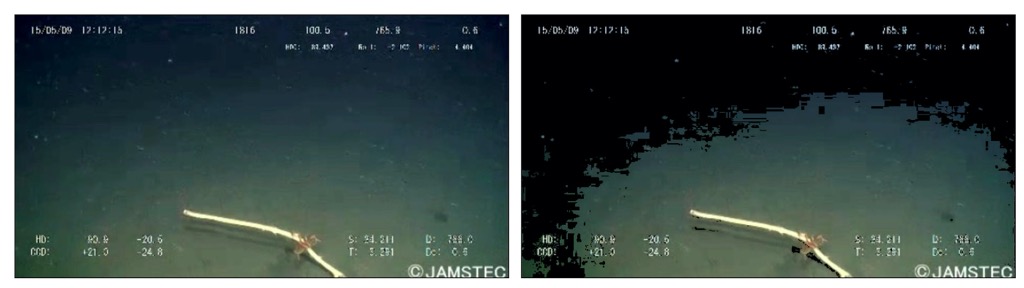} 
    \caption{Background removal. }
    \label{fig:b-mirror.jpg}
\end{figure}

\section{Experiments}
\subsection{Experimental Setup}
%\subsection{Training and Testing }
We conduct our experiments mainly on the TrashCan 1.0 data set \cite {Hong2020TRASHCAN:DEBRISb} with both instance-based and material-based versions as given in Table 1. The data set is split into a train and validation set. This data set is comprised of two versions. Material-version and Instance-version. The total categories in this dataset are three, while 34 are subcategories. For the material version, we have 16 classes, while for the case of the instance version, we have 22 classes. 
We use the state-of-the-art Detectron2 for our work. 
Instead of developing the RetinaNet model \cite{FocalCode} from scratch, we use it to reduce our development time and help increase speed. 
In our problem settings, we use our simple enhancement strategy along with a single-stage detector RetinaNet \cite{FocalCode} as the backbone in Detectron2 \cite{Facebookresearch/detectron2:Tasks.}. Detectron2   implements state of the art object detection  algorithms \cite{{Facebookresearch/detectron2:Tasks.}} 
Out of the two pre-trained weights, \texttt{retinanet\_R\_101\_FPN\_3x.yaml} and \texttt{retinanet\_R\_50\_FPN\_3x.yaml} the former give us better results which give us enhanced accuracy while further keeping the processing time less as compared to the two-stage detectors. 
In our problem settings, we use our simple enhancement strategy along with a single-stage detector RetinaNet \cite{FocalCode} as the backbone in Detectron2 \cite{Facebookresearch/detectron2:Tasks.}. We tune relevant hyperparameters including learning rates, batch size and others.   
 
\subsection{Implementation Details}
We initialize the network under the default settings, keeping RetinaNet \cite{FocalCode} as the baseline for most experimentation. After the image enhancement step, when the image is input to the model, we further experiment and evaluate the performance of FRCNN and RetinaNet \cite{FocalCode} with average precision metrics. Initially, we run the experiments using unprocessed raw data. We then use IEM to preprocess the image before passing it via the channel stabilization module. Then we apply the augmentations and channel sharpening filter to improve the results further. First, we run the set of experiments for the TrashCan1.0 material version and then run it for the TrashCan \cite{Hong2020TRASHCAN:DEBRISb} instance version. After getting the results with Detectron2 \cite{Facebookresearch/detectron2:Tasks.} 
we compare our results by conducting experiments on Deformable Transformer. Deformable DETR is an end-to-end detector based on a sampling-based efficient attention mechanism that makes it much simpler and has faster convergence \cite{Zhu2020DeformableDetection}.
SGD is used to optimize the training loss. Recent analytical research shows that SGD tends to converge towards a flatter minimum in the loss landscape compared to ADAM, which translates to a better generalization performance. In addition, we adjust the learning rate during the experiments to find an optimal balance between bounding box prediction and object classification. We reproduce base results by tuning the hyper-parameters like LR and batch size. We use a batch size of 2 and perform training using a single GPU. %We use a learning rate of o.0025 to 0.00025 
Empirically, we set LR = 0.00025 for training the network with the maximum number of 40k iterations in Detectron2. We also use one GPU for evaluation. By keeping settings same both in training as well as in validation, We obtain the bounding box prediction and individual category results. To produce the final detection, we apply non-maximum suppression (NMS) with a threshold of 0.7 combining the top predictions from all classes. While higher value of NMS helps in giving single , smooth bounding box prediction, careful selection of this is recommended. 
\begin{table*}[]
\centering
\caption{Comparison of the average accuracy (AP) of different object detectors at different IOUs,[AP, AP50, AP75]. Object detection performance for objects including  small size[APS], medium size[APM] and large size[APL]}
\label{tab:my-table}
\begin{tabular}{@{}ccccccc@{}}
\toprule
\textbf{Method} &
  \textbf{AP} &
  \textbf{AP50} &
  \textbf{AP75} &
  \textbf{APS} &
  \textbf{APM} &
  \textbf{APL} \\ \midrule
\multicolumn{7}{c}{Dataset -Material}                                           \\ \midrule
Base   line                    & 29.10  & 51.21  & 27.83  & 28.22  & 30.20  & 40.0   \\
RetinaNet                      & 32.51 & 52.38  & 34.84 & 29.23 & 31.01 & 45.0  \\
Retinanet+IEM &
  \textbf{36.11} &
  \textbf{56.7} &
  \textbf{38.68} &
  \textbf{33.59} &
  \textbf{34.55} &
  \textbf{47.75} \\
FRCNN                          & 32.92  & 50.36 & 37.20  & 28.30 & 31.53 & 41.16 \\
FRCNN + IEM                    & 35.10  & 52.70  & 38.92  & 29.01  & 35.12  & 48.20   \\
Deformable  Transformer        & 34.15  & 57.24  & 36.01  & 29.12  & 33.72  & 49.7   \\
Deformable  Transformer + IEM  & 34.91  & 58.02    & 36.30  & 28.92  & 34.23  & 49.90   \\ \midrule
\multicolumn{7}{c}{Dataset-Instance}                                            \\ \midrule
Baseline                       & 34.54  & 55.41  & 38.10  & 27.62  & 36.21  & 51.4   \\
RetinaNet                      & 41.02 & 60.70  & 45.80  & 38.63  & 43.05 & 62.18  \\
RetinaNet+(IEM) &
  \textbf{44.03} &
  \textbf{63.09} &
  \textbf{48.38} &
  \textbf{40.37} &
  \textbf{46.54} &
  \textbf{64.27} \\
FRCNN                          & 38.51  & 54.05 & 44.03  & 26.02  & 43.03  & 63.07   \\
FRCNN + IEM &33.54 &48.20 &36.88 &24.44 &38.55 &57.80\\
%   \multicolumn{1}{l}{35.12} &
%   \multicolumn{1}{l}{52.07} &
%   \multicolumn{1}{l}{38.93} &
%   \multicolumn{1}{l}{29.06} &
%   \multicolumn{1}{l}{35.22} &
%   \multicolumn{1}{l}{48.25} &  
Deformable   Transformer       & 39.08 & 63.03  & 44.51  & 39.01   & 42.50  & 64.01     \\
Deformable   Transformer + IEM & 39.60  & 63.87  & 44.15  & 38.82  & 41.42  & 62.33   \\ \bottomrule
\end{tabular}
\end{table*}

\subsection{Ablation Study}
We conduct ablation study with Channel Sharpening module, with addition of channel stabilization module. We further try different augmentations to observe their effects in our problem settings. We evaluate all these options using different backbones available in Detectron2 \cite{Facebookresearch/detectron2:Tasks.}. We add these components one by one  for both version of data sets while keeping the same settings. First we take material version of data set and in order to explore the most suitable backbone which give us best results we initially start by using RetinaNet As a backbone with its relevant pre-trained weights, we used \texttt{retinanet\_R\_101\_FPN\_3x.yaml} as well as \texttt{retinanet\_R\_50\_FPN\_3x.yaml} and it came out that the 
 that the one with 101 depth give us good results. So we keep it same  for the rest of the experiments. Then we pass the image through image pre-processing /enhancement module and then it pass through channel stabilization module. It is clear from the Table 2 that image enhancement with CSM give us 7 \% improvement in terms of accuracy. The result is consistent in terms of bounding box detection, for large size and medium size objects, while it shows  quite a significant improvement for small size objects which show 5.39 \% increase in accuracy. Keeping the same settings we only change the back bone with FRCNN and get the required results. These show 6\%
 improvement in AP values of bounding box detection where as for the small size object detection it shows 1\% improvement. 
 After extensive testing for different backbones for material version of data set we repeat the above procedure for Instance version of data set for all back bones. For Instance version for RetinaNet \cite{FocalCode} we observe absolute gain of  9.53\% increase in bounding box predictions(AP) as can be seen from Table 1, whereas for small object detection we see 12.7 \% increase which shows the biggest increase as compared to the material version of the dataset. For FRCNN \cite{Ren2015FasterNetworks} Table 1 shows that it give us 4 \% increase in detection of small objects and for bounding box detection it shows 1.4\% improvement. Hence over all our method shows the increase in detection accuracies, nonetheless RetinaNet with image enhancement and channel stabilization outperform all others. Figure 7 and 8 shows the performance comparison for individual class detection between FRCNN and RetinaNet with reference to base line for material version and instance version respectively. We can see in most of the cases RetinaNet supercedes in performance except animal eel, trash fabric ,trash fishing gear and plants for instance version and RetinaNet gives better performance in all others eleven classes, while for the case of material version apart from one class trash wreckage its RetinaNet is performs well in other classes.
%  Deformable DETR \cite{Zhu2020DeformableDetection} is an end-to-end object detector, which is efficient and fast-converging. It enables us to explore more interesting and practical variants of end-to-end object detectors. 
%At the core of Deformable DETR are the (multi-scale) deformable attention modules, which is an efficient attention mechanism in processing image feature maps.
 We compare our results with DETR in order to evaluate performance of our detector and as can be seen in Table 1 and it give us 9.5\% improvement in accuracy with reference to object bounding box detection where as for small size detection it give us 11\% increase in accuracy for instance version of dataset which is less as compared to the accuracy we obtain from RetinaNet with same number of epochs. For material version case deformable transformer give us 5.4 \% increase in accuracy while for small size detection it give us almost .7\% increase in accuracy which is again less as compared with our previous results. One reason for this  may be limited number of samples as well as different combination of augmentation methods. We also train our model without any pre-trained weights and it give  better results as compared to the ones with pre-trained weights. We observe that by using this simple image enhancement method with one stage detector can give us much better results with less processing and computation required. Though better combination of augmentation techniques along with  image enhancement method (IEM), with state of the art methods can give us better results.

% Please add the following required packages to your document preamble:
% \usepackage{graphicx}

% Please add the following required packages to your document preamble:
% \usepackage{graphicx}
% Please add the following required packages to your document preamble:
% \usepackage{graphicx}

% \begin{table*}[]
% \caption{Comparison of the average accuracy (AP) of  our object detector  with  TRASHCAN data set.}
% \centering
% \resizebox{\textwidth}{!}{%
% Please add the following required packages to your document preamble:
% \usepackage{graphicx}
% \usepackage[table,xcdraw]{xcolor}
% If you use beamer only pass "xcolor=table" option, i.e. \documentclass[xcolor=table]{beamer}

% Please add the following required packages to your document preamble:
% \usepackage{booktabs}
% \usepackage{graphicx}
% \usepackage[table,xcdraw]{xcolor}
% If you use beamer only pass "xcolor=table" option, i.e. \documentclass[xcolor=table]{beamer}

%\resizebox{\textwidth}{!}{%
% Please add the following required packages to your document preamble:
% \usepackage{booktabs}
% \usepackage{graphicx}

% Please add the following required packages to your document preamble:
% \usepackage{booktabs}
% \usepackage{graphicx}

% Please add the following required packages to your document preamble:
% \usepackage{booktabs}
% \usepackage{graphicx}
\begin{table*}[]
% Please add the following required packages to your document preamble:
% \usepackage{booktabs}

\centering
\caption{Comparison of the average accuracy (AP) of proposed object detector at different IOUs,[AP, AP50, AP75]. Object detection performance for objects including  small size[APS], medium size[APM] and large size[APL] is also given}
\label{tab:my-table}
\begin{tabular}{@{}ccccccc@{}}
\toprule
\textbf{Method}                                             & \textbf{AP}    & \textbf{AP50}  & \textbf{AP75}  & \textbf{APS}   & \textbf{APM}   & \textbf{APL}   \\ \midrule
\multicolumn{7}{c}{Dataset -Material}                                                                                                                             \\ \midrule
Base   line                                                 & 29.16           & 51.20           & 27.8 2          & 28.21           & 30.23           & 40.0           \\
RetinaNet                                                   & 32.52          & 52.32           & 34.82          & 29.25          & 31.01          & 45.04          \\
\multicolumn{1}{l}{RetinaNet+Channel Stablization+Guassian} & 32.90           & 52.89          & 34.51          & 32.09          & 31.62          & 42.56          \\
\multicolumn{1}{l}{RetinaNet+CSM+IEM+Sharpening Filter} & 35.98     &56.10 &37.60.  &32.93 &33.53 &47.30
%   \multicolumn{1}{l}{35.98} &
%   \multicolumn{1}{l}{56.10} &
%   \multicolumn{1}{l}{37.60} &
%   \multicolumn{1}{l}{32.93} &
%   \multicolumn{1}{l}{33.53} &
%   \multicolumn{1}{l}{47.30} 
  \\
RetinaNet + Channel Stablization + IEM                         & \textbf{36.11} & \textbf{56.7}  & \textbf{38.68} & \textbf{33.59} & \textbf{34.55} & \textbf{47.75} \\ \midrule
\multicolumn{7}{c}{Dataset-Instance}                                                                                                                              \\ \midrule
Baseline                                                    & 34.52           & 55.42           & 38.13           & 27.61           & 36.20           & 51.4           \\
RetinaNet                                                   & 41.02          & 60.72          & 45.82           & 38.61           & 43.03          & 62.18          \\
\multicolumn{1}{l}{RetinaNet+Channel Stablization+Guassian} &41.02 &61.02 &46.50 &38.90 &44.10 &63.05
%   \multicolumn{1}{l}{41.92} &
%   \multicolumn{1}{l}{61.02} &
%   \multicolumn{1}{l}{46.5} &
%   \multicolumn{1}{l}{38.90} &
%   \multicolumn{1}{l}{44.10} &
%   \multicolumn{1}{l}{63.05} 
\\
\multicolumn{1}{l}{RetinaNet+CSM+IEM+Sharpening Filter} &39.54 &57.85 &42.65 &32.44 &43.14 &64.07
%   \multicolumn{1}{l}{39.54} &
%   \multicolumn{1}{l}{57.85} &
%   \multicolumn{1}{l}{42.65} &
%   \multicolumn{1}{l}{32.44} &
%   \multicolumn{1}{l}{43.14} &
%   \multicolumn{1}{l}{64.07} 
\\
RetinaNet+ImageEnhancement Model(IEM)                      & \textbf{44.03} & \textbf{63.09} & \textbf{48.38} & \textbf{40.37} & \textbf{46.54} & \textbf{64.27} \\ \bottomrule
\end{tabular}

\end{table*}
\subsection{State-of-the-art Comparison}
\subsubsection{Quantitative Results }

It is clear from the Table 1 that with our simplest image enhancement model and improving the color dominance  with channel stabilized method, we get improvement in performance in  bounding box detection and  small size object detection along with medium and big size objects. We obtain this increase for both  material version as well as Instance version of TrashCan 1.0 data set \cite{Hong2020TRASHCAN:DEBRISb}. For quantitative comparison we use mAP values [AP, APS, APL] for bounding box detection as well as for multi scale prediction so that we can have fair comparison across the varied range of  classes. It shows that our approach performs quite well as compared to the baseline. We relate this improvement to enhancement strategy with sharpened edges. The sharpening filter enhances the edges and channel stabilization with reduces the color haze and help us achieve better prediction. The back ground removal methods perform well in some cases while along with image enhancement it does not give good results. For quantitative comparison we get the results of Faster R-CNN, RetinaNet, Deformable transformer to validate our findings.
The experimental results further confirm that some augmentations introduced along with channel stabilization module helps getting quite good accuracy as compared to others. Table 2 shows the performance enhancment with addition of IEM blocks like Channel Stabilization module (CSM), Sharpening filter and augmentations respectively.
To observe the individual class detection performance, we compare the individual class detection performance of our method, for both material version and instance version of data set, against the baseline as given in Figure 5 and 6.
\begin{figure}[h]
   \centering
    \includegraphics[scale=0.20]{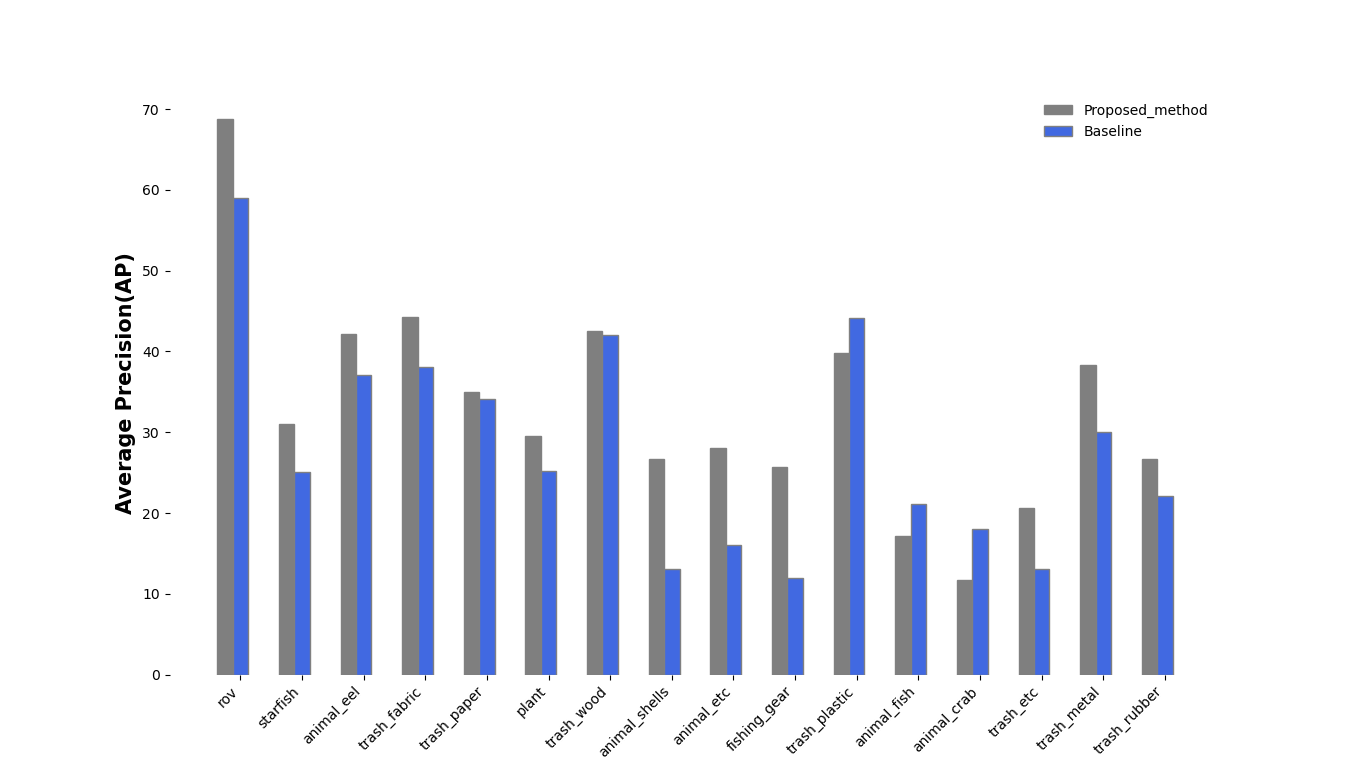} 
    \caption{Individual Class Detection for Retina Net - Material-version. }
    \label{fig:b-mirror.jpg}
\end{figure}
\begin{figure}[h]
   \centering
    \includegraphics[scale=0.16]{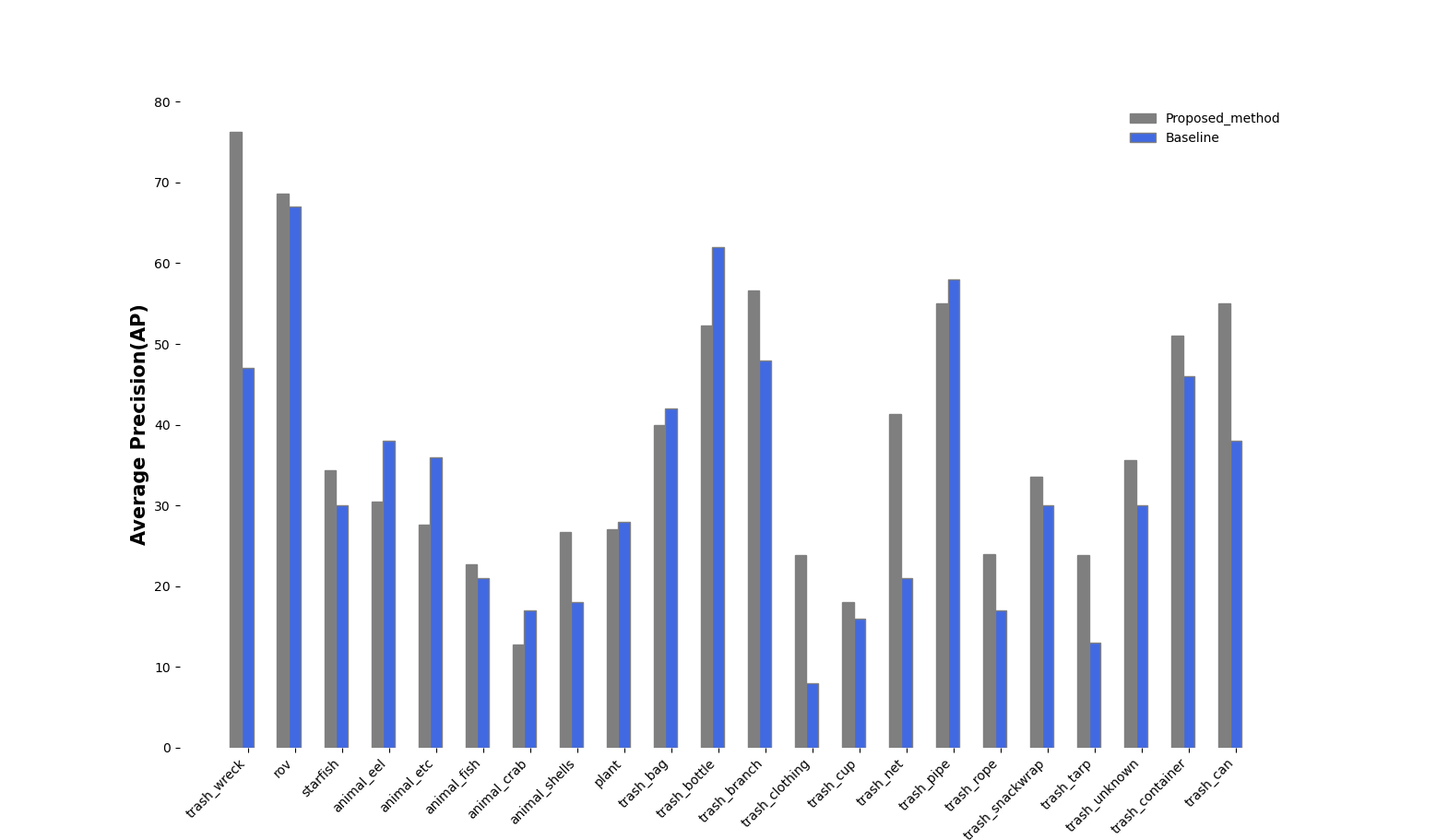} 
    \caption{Individual Class Detection for Retina Net - Instance-version.  }
    \label{fig:mirror.jpg}
\end{figure}
\begin{figure}[h]
   \centering
    \includegraphics[scale=0.35]{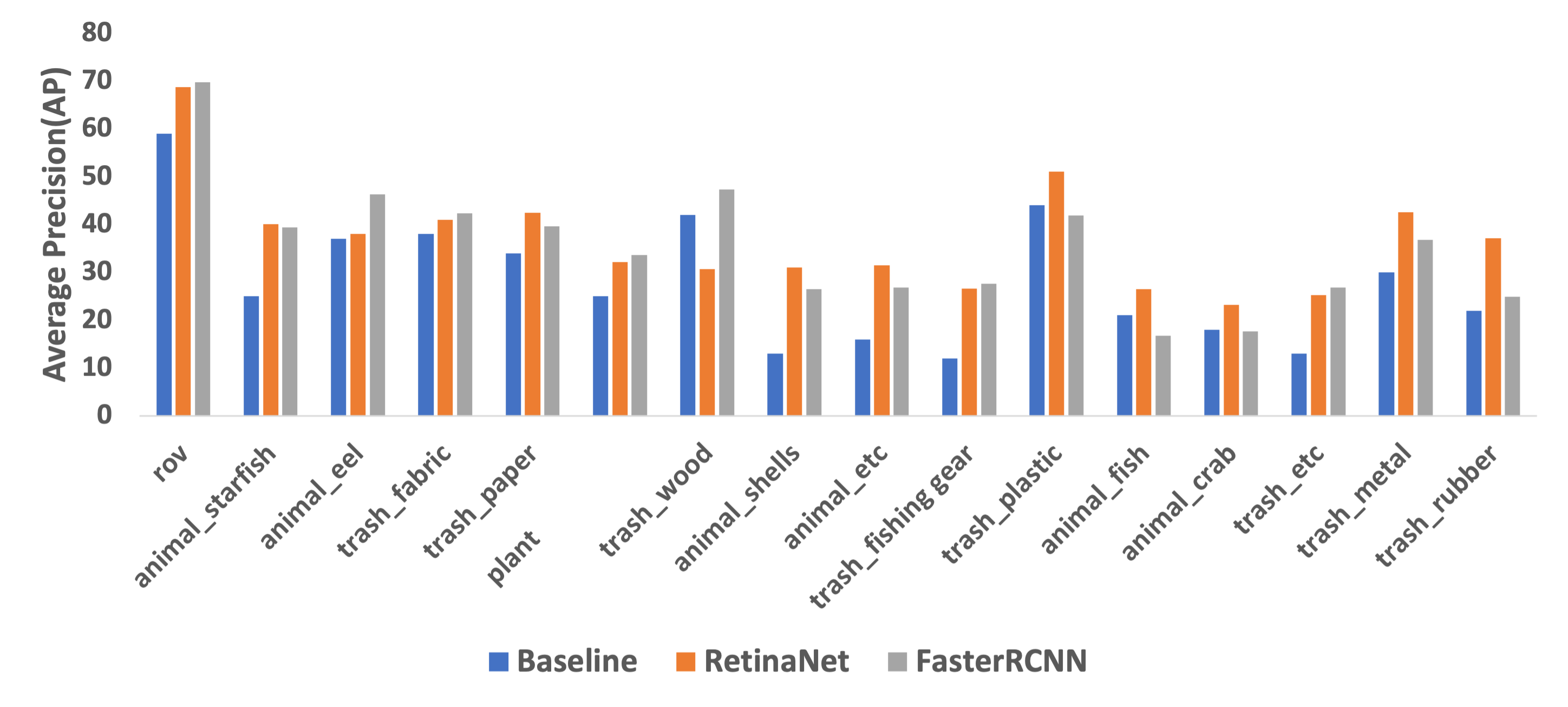} 
    \caption{Backbone Detection Comparison:Individual Class  Average Precision - Material - version. }
    \label{fig:mirror.jpg}
\end{figure}
\begin{figure}[h]
   \centering
    \includegraphics[scale=0.40]{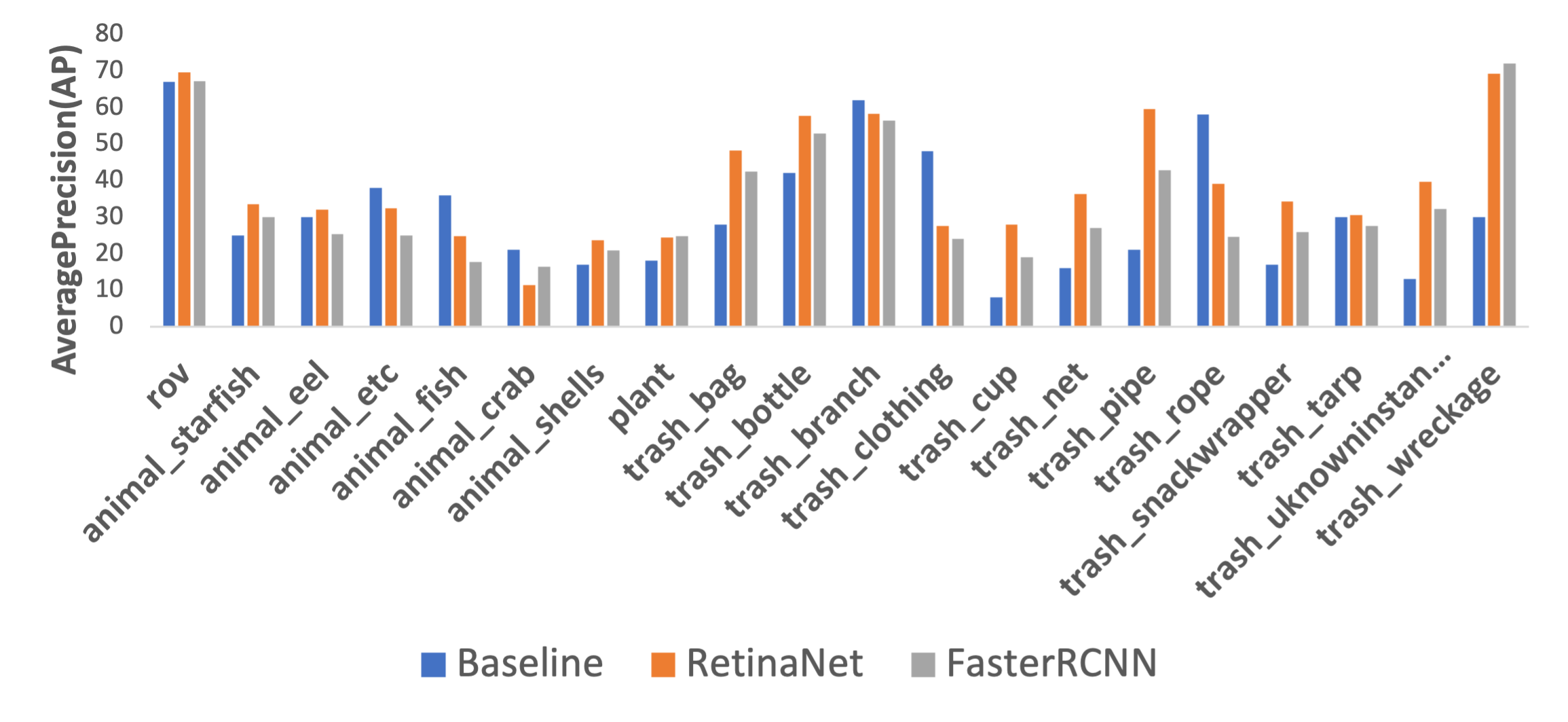} 
    \caption{Backbone Detection Comparison:Individual Class  Average Precision - Instance - version. }
    \label{fig:backbone.jpg}
\end{figure}\\
We explore the detection performance for individual classes and we show good improvement in detection accuracy across all the classes except animal-crab for the case of material version and trash rope in the case of instance version. The individual class comparison is given in Figure 5 and Figure 6. We also observe the detection of individual classes across Top2 backbones in comparison with the baseline. The result in Figure 7 and Figure 8 how RetinaNet outperforms for individual detection of classes. For some classes like wood and eel our method with RetinaNet is not the best one but still it exceeds the base line with good margin, for all the other classes it gives best results. 
% The effects of  channel stabilization only as well as channel stabilization with image enhancement module is given in Figure 7 and Figure 8.
\subsubsection{Qualitative Results}
Our method works in most of the cases using this simpler approach, yet it face some difficulty
for prediction where colour of object is very dull. We show that our approach works well for different size objects present in the picture and we get prediction accuracy increase for multiple objects together. The scheme shows good performance for the cases where we have only single object present or multiple objects present. Further we see with image enhancement and channel stabilization we remove the colour dominance of one particular colour which is blue in our case and give us samples with better colours and detection. We show few challenging qualitative prediction samples in Figure 9 and Figure 10. For the cases where there is less blur and image colours are not bright model performs exceptionally well.

\begin{figure}[h]
   \centering
    \includegraphics[scale=0.50]{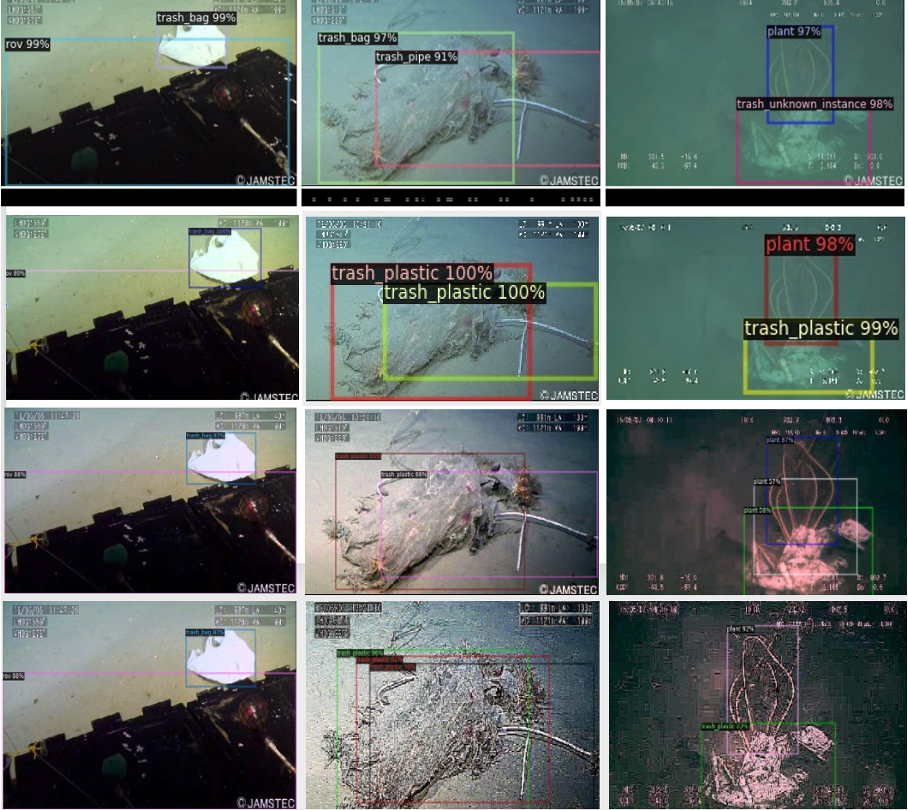} 
    
    \caption{Few sample detections using our approach on TrashCan 1.0 dataset-material version: Each consecutive row shows  base line image, image with enhanced module and image with enhancement and channel stabilized module respectively. We can see improvement at each stage.}
     \label{fig:pic1.jpg}
\end{figure}
\begin{figure}[h]
   \centering
    \includegraphics[scale=0.50]{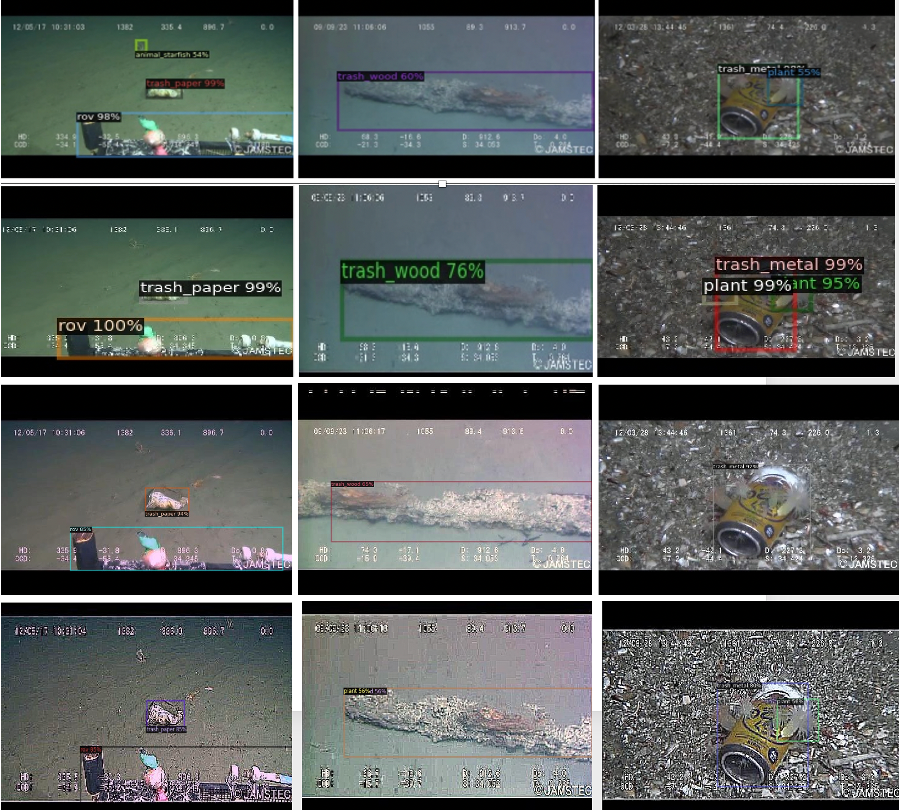} 
    
    \caption{Few sample detections using our approach on TrashCan 1.0 dataset-Instance version: Each consecutive row shows  base line image, image with enhanced module and image with enhancement and channel stabilized module respectively. We can see improvement at each stage.  }
     \label{fig7:pic-inst}
\end{figure}

% Please add the following required packages to your document preamble:
% \usepackage{booktabs}

We compare our results with existing state-of-the-art methods on this dataset as given in Table 3. It shows clearly for all metrics use we see consistent improvement with 6.11\% absolute gain in AP for material version of dataset used and almost 12 \% increase for instance version of the dataset. These results  give the comparison  in Table 3 both for material version and instance version of the data set. 
\begin{table}[]
\centering
\resizebox{\columnwidth}{!}
\
\caption{Performance evaluation of our approach with other state of the art methods}
\label{tab:my-table}
\begin{tabular}{@{}ccccccc@{}}
\toprule
\textbf{SOTA Methods} & \textbf{AP}    & \textbf{AP50}  & \textbf{AP75}  & \textbf{APS}   & \textbf{APM}   & \textbf{APL}   \\ \midrule
\multicolumn{7}{c}{Dataset-Material}                                                                                        \\ \midrule
Mask R-CNN            & 30.0          & 51.2          & 27.8           &28.2          & 30.2          & 40.0           \\
Faster R-CNN         & 34.45         & 55.40         & 38.60          &27.60        & 36.20         & 51.4         \\
\textbf{Our Approach} & 
  \multicolumn{1}{l}{\textbf{36.11}} &
  \multicolumn{1}{l}{\textbf{56.7}} &
  \multicolumn{1}{l}{\textbf{38.68}} &
  \multicolumn{1}{l}{\textbf{33.59}} &
  \multicolumn{1}{l}{\textbf{34.55}} &
  \multicolumn{1}{l}{\textbf{47.75}} \\ \midrule
\multicolumn{7}{c}{Dataset-Instance}                                                                                        \\ \midrule
Mask R-CNN   & 32.52          & 52.3           & 34.82          & 29.25          & 31.01          & 45.04          \\
Faster R-CNN & 32.9           & 50.35          & 37.2           & 28.32          & 31.54          & 41.16         \\
\textbf{Our Approach}         & \textbf{44.03} & \textbf{63.09} & \textbf{48.38} & \textbf{40.37} & \textbf{46.54} & \textbf{64.27} \\ \bottomrule
\end{tabular}
\end{table}
\subsection{Domain Generalization }
% Please add the following required packages to your document preamble:
% Please add the following required packages to your document preamble:
% \usepackage{booktabs}
\begin{table}[]
\centering
\caption{ Performance evaluation of our approach on other data sets.}
\label{tab:my-table}
\begin{tabular}{@{}ccccccc@{}}
\toprule
\textbf{Dataset} & \textbf{AP} & \textbf{AP50} & \textbf{AP75} & \textbf{APS} & \textbf{APM} & \textbf{APL} \\ \midrule
UAVVaste        & 76.604      & 95.89         & 87.34         & 53.21        & 79.64        & 86.53        \\
%-19         & 88.86       & 99.56         & 98.86         & 68.06        & 85.50        & 90.64        \\
TACO-60         & 15.2        & 18.10         & 17.42         & 2.03         & 7.83         & 16.93        \\
%-19         & 88.86       & 99.56         & 98.86         & 68.06        & 85.50        & 90.64        \\
\multicolumn{1}{l}{UODD} &
  \multicolumn{1}{l}{20.94} &
  \multicolumn{1}{l}{45.6} &
  \multicolumn{1}{l}{16.35} &
  \multicolumn{1}{l}{13.45} &
  \multicolumn{1}{l}{22.47} &
  \multicolumn{1}{l}{22.53} \\ \bottomrule
\end{tabular}
\end{table}
To further investigate the performance on other data sets , we benchmarked  our result on  UAVVaste, TACO, and UODD. Table 4 shows these results for Faster R-CNN as well as Mask R-CNN \cite{He2017MaskR-CNN}, and Fig 11 shows sample detections for other data sets.
We apply our techniques on various data sets and show improved performance as give in Table 4. we give few qualitative samples for different dataset's results. We compare existing results on these datasets where EffecientNet on UAVVaste give us AP50 of 79.90 where we get 95.89 as given in Table 3. For comparison with other methods we need results for individual classes as well as average precision values for different scales which are not available in the literature till now. Right combination of augmentations methods with background removal technique may improve performance for night images.  
%\begin{table}[htbp]
%\ContinuedFloat
%\begin{subtable}{\linewidth}
%\centering
%\caption{Results for TACO Dataset}
%\begin{tabular}{@{}c|cccccc@{}}
%\toprule

%\textbf{Method} & \textbf{AP}  & \textbf{AP50}  & \textbf{AP75}  & \textbf{APS}  & \textbf{APM}  & \textbf{APL}  \\ \midrule
%RetinaNet           & 15.2          & 18.8          & 17.4         & 1.34          & 7.8         & 16.9         \\
%FRCNN        & {15.2} & 19.2         & 17.14         & 0.6         & 8.0          & 19.3          \\
 %\\ \bottomrule
%\end{tabular}
%\label{tab:resultssecondopt}
%\end{subtable}

%\label{tab:results}
%\end{table}

\begin{figure}[h]
   \centering
    \includegraphics[scale=0.25]{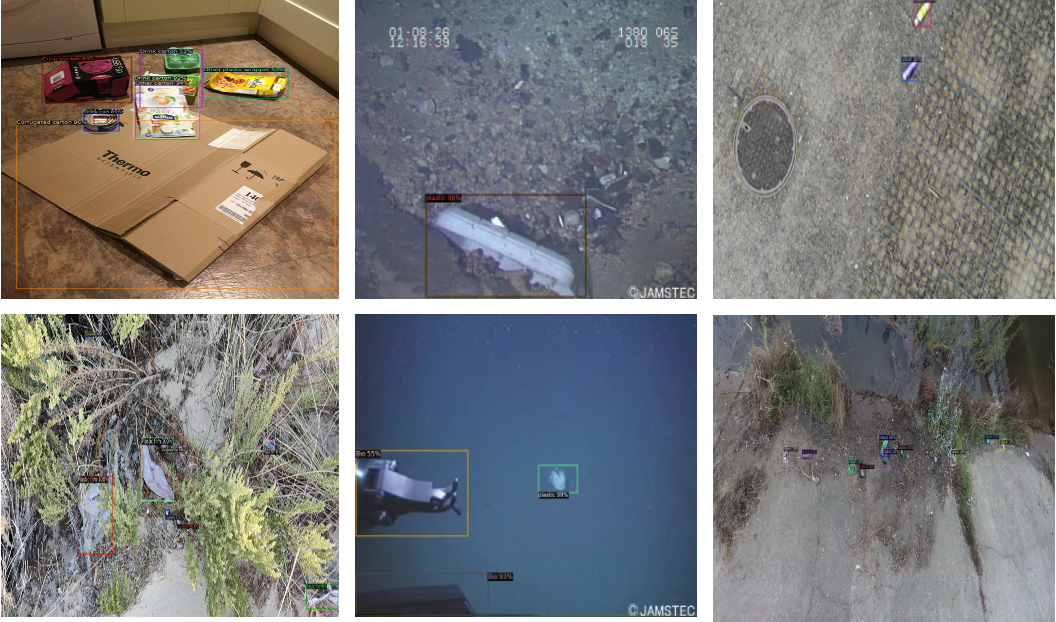} 
    \caption{ Qualitative samples o TACO, Trash-ICRA19, UAVVaste, respectively from left to right column-wise }
    \label{fig:uav.png}
\end{figure}
.

\section{Conclusion}
As underwater object detection is a significant real-world problem, we introduce image enhancement methods that significantly improve detection accuracy by using a simplified image enhancement method named Image Enhancement Module (IEM) and a novel channel stabilization module. These strategies remove the color haze from images to improve their visibility. We propose that raw unprocessed underwater images for objection detection are insufficient in achieving good results as the blue color dominates most underwater images due to the physical phenomenon of light scattering and absorption. The IEM consists of augmentations as well as a sharpening filter before being passed to the detector. We explored the different state-of-the-art detection methods and compared their performances in the underwater detection task. We show that by applying an effortless image enhancement technique, we exploit the best capabilities of a single-stage detecto.The final results indicate that employing RetinaNet with X101-FPN as the base model and using our simple enhancement method produces better prediction results for underwater detection overall, especially for small-scale objects. 
% Therefore, single-stage detector with our simple enhancement modules reduces the computational cost in real time while giving better prediction results.

% conference papers do not normally have an appendix

% use section* for acknowledgement
% \section*{Acknowledgment}

% The authors would like to thank...

% trigger a \newpage just before the given reference
% number - used to balance the columns on the last page
% adjust value as needed - may need to be readjusted if
% the document is modified later
%\IEEEtriggeratref{8}
% The "triggered" command can be changed if desired:
%\IEEEtriggercmd{\enlargethispage{-5in}}

% references section

% can use a bibliography generated by BibTeX as a .bbl file
% BibTeX documentation can be easily obtained at:
% http://www.ctan.org/tex-archive/biblio/bibtex/contrib/doc/
% The IEEEtran BibTeX style support page is at:
% http://www.michaelshell.org/tex/ieeetran/bibtex/
\bibliographystyle{IEEEtran}
% argument is your BibTeX string definitions and bibliography database(s)
\bibliography{references.bib}
%
% <OR> manually copy in the resultant .bbl file
% set second argument of \begin to the number of references
% (used to reserve space for the reference number labels box)
% \begin{thebibliography}{10}

% %\bibitem{IEEEhowto:kopka}

% \end{thebibliography}

% that's all folks
\end{document}